\documentclass[biblatex, english]{lni}
\usepackage{booktabs}
\usepackage[]{blindtext}
\graphicspath{{figures/}}
\usepackage{multirow}
\usepackage{subcaption}
\usepackage{makecell}
\begin{document}
%%% Mehrere Autoren werden durch \and voneinander getrennt.
%%% Die Fußnote enthält die Adresse sowie eine E-Mail-Adresse.
%%% Das optionale Argument (sofern angegeben) wird für die Kopfzeile verwendet.
\title[Segmenting Wood Rot]{Segmenting Wood Rot using Computer Vision Models}
%%%\subtitle{Untertitel / Subtitle} % falls benötigt
%\author[1]{BLIND REVIEW}{BLIND@REVIEW}{}
%\author[1]{BLIND REVIEW}{BLIND@REVIEW}{}
%\author[1]{BLIND REVIEW}{}{}%
\author[1]{Roland Kammerbauer}{kammerbauerro76348@th-nuernberg.de}{}
\author[1]{Thomas H. Schmitt}{thomas.schmitt@th-nuernberg.de}{}
\author[1]{Tobias Bocklet}{}{}%
%\affil[1]{BLIND REVIEW\\BLIND REVIEW\\
%BLIND REVIEW\\BLIND REVIEW\\BLIND REVIEW}
\affil[1]{Technische Hochschule Nürnberg Georg Simon Ohm\\Center for Artificial Intelligence\\ Keßlerplatz 12\\90489 Nuremberg\\Germany}
\maketitle
\begin{abstract}
In the woodworking industry, a huge amount of effort has to be invested into the initial quality assessment of the raw material.
In this study we present an AI model to detect, quantify and localize defects on wooden logs. This model aims to both automate the quality control process and provide a more consistent and reliable quality assessment. 
For this purpose a dataset of $1424$ sample images of wood logs is created. A total of 5 annotators possessing different levels of expertise is involved in dataset creation.
An inter-annotator agreement analysis is conducted to analyze the impact of expertise on the annotation task and to highlight subjective differences in annotator judgement.
We explore, train and fine-tune the state-of-the-art InternImage and ONE-PEACE architectures for semantic segmentation. 
The best model created achieves an average IoU of $0.71$, and shows detection and quantification capabilities close to the human annotators.
\end{abstract}
\section{Introduction}
\label{sec:intro}
In recent years, machine learning has proven to be an essential tool in automating quality control in industrial processes,
as the inference of even the most complex machine learning models is magnitudes faster than manual assessment by humans. Furthermore, as machine learning models operate predictably and consistently regardless of time and day, they prove to be invaluable for objective quality control.
One industry which can benefit enormously from automated quality control, which has maintained its relevance since the earliest stages of humanity, is the woodworking and lumbering industry.
Since wood as a natural product possesses hugely varying quality, judging, sorting and properly utilizing wood based on its quality is a huge effort throughout its entire processing.
One of the most important quality measurements for wooden logs is the presence and amount of wood decay, as logs containing rot, depending on its magnitude, can not be used in many production scenarios. 
Therefore, it is paramount to sort out unsuitable logs as early as possible.
To properly assess the quality of wood logs, skilled experts are required to manually inspect each log and decide if it can be used in further processing, which is both time and labor-intensive.
In this study we present an application of computer vision models to detect and segment rot and other wood defects in images of log crosscuts.
The main contributions of this study are:
\begin{enumerate}
\item Collection and annotation of a dataset to train computer vision models for our highly specific task.
\item Training and comparing several computer vision models and training setups on our segmentation task.
\item Performing an inter-annotator agreement analysis between expert annotators, layman annotators and our best performing model.
\end{enumerate}
\subsection{Related Works}
\label{sec:related_work}
Application studies \cite{doi:10.1139/cjfr-2020-0340, dac3411a39e9484e9284ea2ea969adf6, s19071579, HE2020107357, hrcak278445} use various approaches to detect or classify rot or other defects at different stages of the wood processing process.
\cite{doi:10.1139/cjfr-2020-0340} examines invasive and non-invasive methods to classify whether a still standing tree is rotten or not.
\cite{dac3411a39e9484e9284ea2ea969adf6} automatically determines whether a felled pine log contains rot using computer vision models, while \cite{s19071579} categorizes the severity of the rot into distinct rot severity classes.
\cite{HE2020107357} uses computer vision models to detect defects other than rot in already processed wood products.
\cite{hrcak278445} uses computer vision models to distinguish between different types of rot in wood already installed in buildings.
In contrast to these, this project aims to not only classify, but localize and quantify a variety of defects on cut wooden logs.
%
%\newline
\newline
When inspecting the available images of wood logs, it becomes apparent that log defects exhibit very few strongly distinguishing features.
This mirrors the challenges faced in numerous applications of computer vision in medical imaging, particularly in the analysis of histology and histopathology, in which image areas with only minor differences to the surrounding materials have to be analyzed.
\cite{10.1007/978-3-031-51026-7_2} utilizes Vision Transformers (ViT) \cite{dosovitskiy2021an} to classify breast cancer in histology images, while \cite{KER2019239} uses the Inception V3 model \cite{7298594} on both images of brain and breast tissue.
\cite{ScheiklLaschewski2020} trains segmentation models on images of organs and tissues as a prerequisite for context-aware assistance in cognitive robotics for laparoscopic surgery.
\section{Data}
\begin{figure}
	\centering % <-- added
	\begin{subfigure}{0.33\linewidth}
		\includegraphics[width=\linewidth, trim={0cm 0 0cm 0},clip]{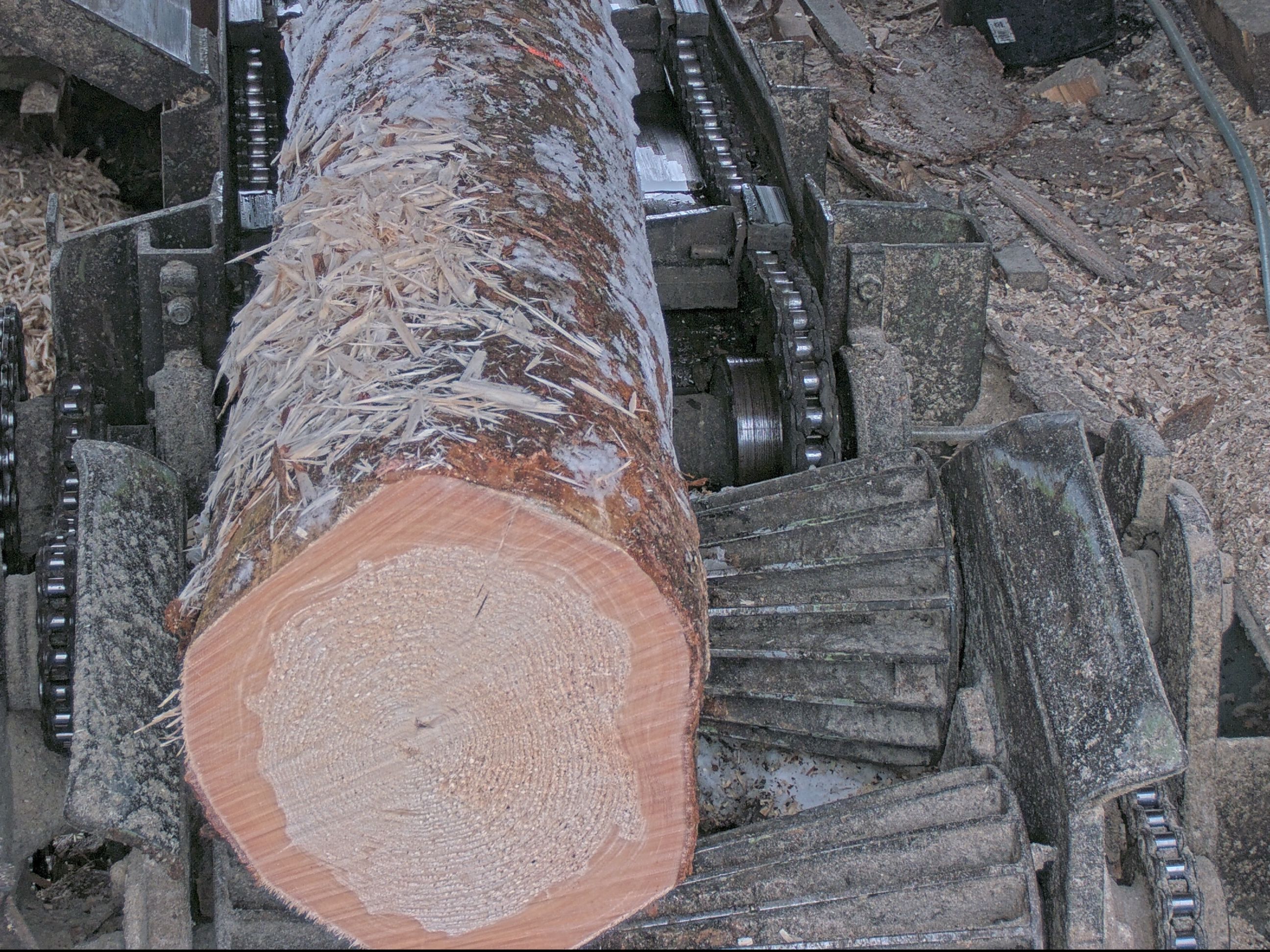}
		\label{fig:examples_1}
	\end{subfigure}\hfil % <-- added
	\begin{subfigure}{0.33\linewidth}
		\includegraphics[width=\linewidth, trim={0cm 0 0cm 0},clip]{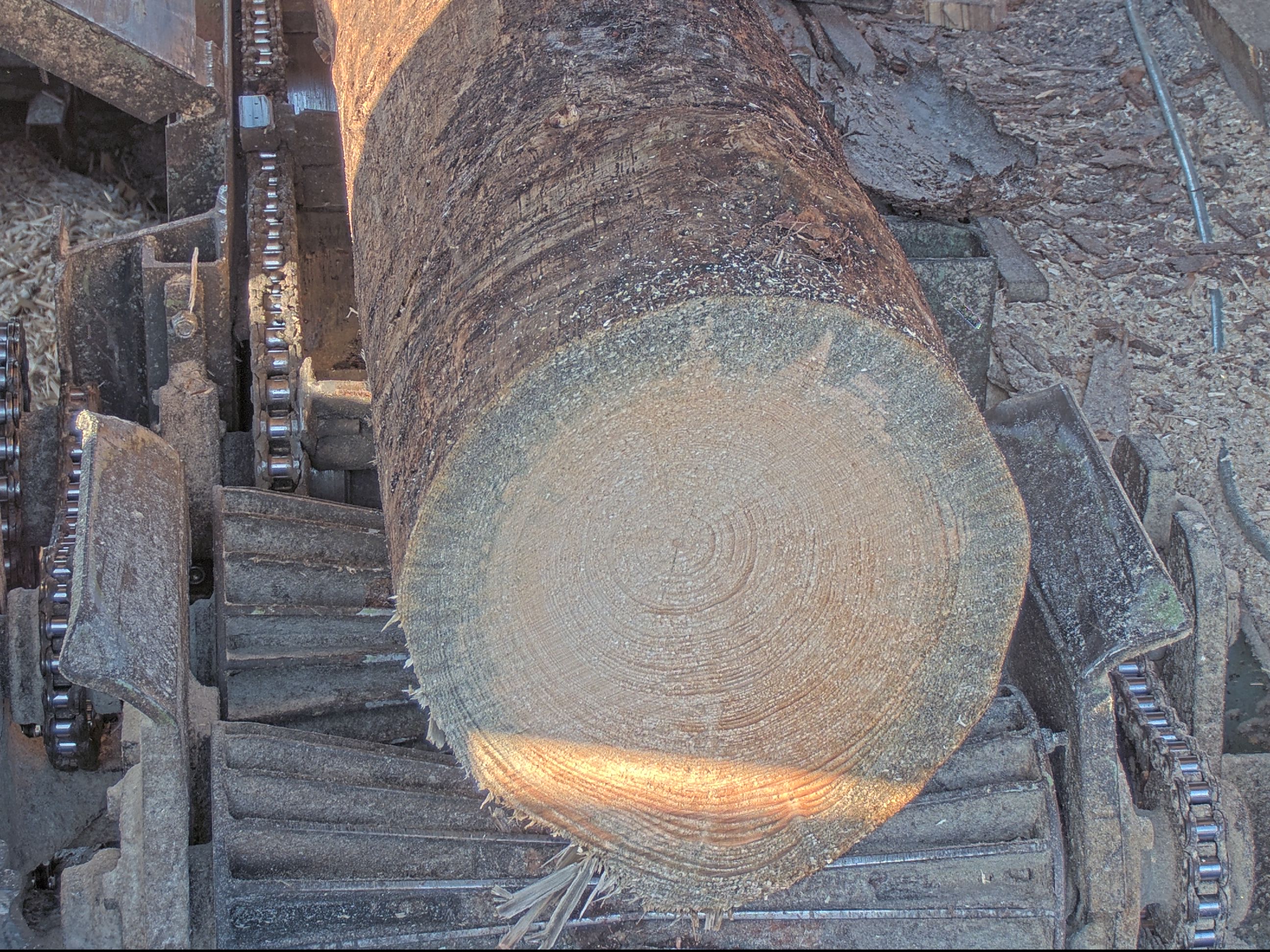}
		\label{fig:examples_2}
	\end{subfigure}\hfil % <-- added
	\begin{subfigure}{0.33\linewidth}
		\includegraphics[width=\linewidth, trim={0cm 0 0cm 0},clip]{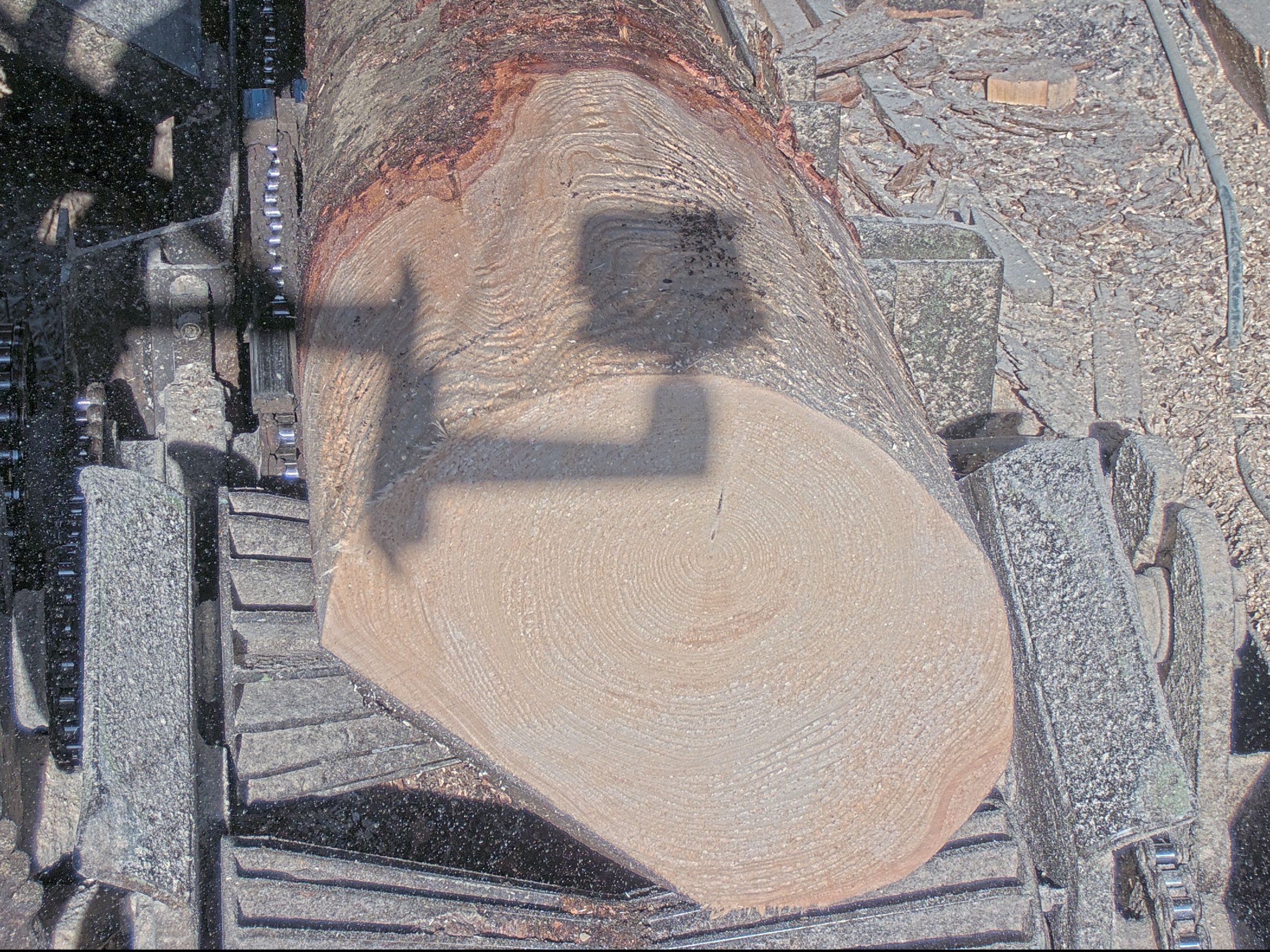}
		\label{fig:examples_3}
	\end{subfigure}
	\caption{A selection of log crosscut images from our dataset, exemplifying the differences in position, weather, and lighting}
	\label{fig:examples}
\end{figure}

To train a computer vision model on our segmentation task we compile a dataset of crosscut images, comprising \(1424\) images in cooperation with a local sawmill.
Images were captured using a statically mounted camera on the conveyor frame, capturing images at a resolution of \(2592px\times1944px\) with a \(4:3\) aspect ratio.
The crosscuts are re-cut before the images are taken as an initial step in the wood processing to ensure a clean, uniform, and straight crosscut.
We limit our dataset to images featuring spruce trees explicitly.
While this narrows down the range of possible defect classes, as certain classes of infestation and pests only attack specific classes of wood \cite{LWFFichte}, it also allows for the segmentation task to focus more strongly on detecting the most prevalent wood defects.
The image background remains relatively constant within the acquired images, although the exact log position, lighting, time of day and prevailing season vary within our dataset due to the real-life variation in conditions when capturing images.
Some example images from the dataset are shown in Fig. \ref{fig:examples}.
%
%\newline
\newline
We manually annotated our dataset using LabelStudio \cite{LabelStudio}.
To denote all relevant image features, \emph{Background}, \emph{Crosscut} and the five defect classes \emph{Rot}, \emph{Rot(maybe)}, \emph{Pressure Wood} (wood where the density and structure of the material is different to regular wood due to external influences), \emph{Discoloration}, and \emph{Ingrowth/Crack} are used in annotation.
\emph{Rot(maybe)} serves as a defect metaclass. Annotators were instructed to mark areas for which they were unsure if they truly represented \emph{Rot} as \emph{Rot(maybe)}.
The images are automatically pre-annotated with \emph{Crosscut} class annotations derived from the biggest object found by a preliminary Segment Anything Model \cite{SAM}.
While our annotators reported that they periodically needed to modify or fully redo the \emph{Crosscut} annotation, this still significantly reduced the required annotation effort.
In total, five different annotators \textit{A} to \textit{E} participated in the image annotation process. Of these five, \textit{A} and \textit{B} are deemed experts at the recognition of wood defects, \textit{C} to \textit{E} are considered laymen.
Before the annotation process, images were divided into multiple subsets, shown in Tab. \ref{tab:datasets}.
\begin{table}
    \begin{center}
        %\scriptsize
        \renewcommand{\arraystretch}{1}
        \begin{tabular}{lrrrrrrr}      
            \toprule
            Set Name & \# Images & CC & R & R(m) & PW & DC & IC \\
            \midrule
            Full Set & 1424 & 85.84 & 4.79 & 2.08 & 2.57 & 3.88 & 0.84 \\
            examples & 58 & 79.81 & 13.46 & 0.38 & 0.94 & 4.59 & 0.81 \\
            warmup & 50 & 89.28 & 5.49 & 0.74 & 0.89 & 1.85 & 1.74 \\
            data & 1316 & 85.97 & 4.38 & 2.21 & 2.71 & 3.93 & 0.81 \\
            \bottomrule
            \multicolumn{8}{l}{Classes BG: background, CC: Crosscut, R: Rot, R(m): Rot(maybe)} \\
            \multicolumn{8}{l}{PW: PressureWood, DC: Discoloration, IC: Ingrowth/Crack}
        \end{tabular}        
        \caption{Dataset subsets with mean distribution of classes w.o. background in percent of the total crosscut area}
        \label{tab:datasets}
    \end{center}
\end{table}
The \textit{examples} subset was annotated by expert annotator \textit{A}, and served as an annotation guide for the laymen annotators (\textit{C} to \textit{E}).
Annotations for the \textit{warmup} subset were created by all annotators, and serve as a baseline for the inter-annotator agreement analysis in Section \ref{sec:sec:inter_rater}.
The \textit{data} subset comprises the remaining images used for model training and evaluation and was annotated by annotators \textit{C} to \textit{E}. 
Since these are layman annotators, their annotations for the \emph{data} subset were checked and subsequently revised by expert annotator \textit{B}.
In total, expert annotator \emph{B} created revised annotations for \(33.7\% \) of the \emph{data} subset.
Tab. \ref{tab:datasets} further shows the average defect class area normalized by the sample-wise crosscut area.
This takes the respective log sizes into account, and therefore closely represents the actual prevalence and amount of the defects.
Unsurprisingly, the defect classes constitute a minority of the normalized image area, with the \emph{Crosscut} class dominating the images.
For initial preprocessing of the annotations, all defect annotations outside of the \emph{Crosscut} area are removed from the ground truth.
As we only consider defects occurring within the crosscut of the logs, any annotations outside of the crosscut are interpreted as unwanted artifacts.
Given that semantic segmentation models operate on disjoint segmentation masks, we enforce a maximum of one class per pixel using a hierarchical approach.
The importance of each class correlates with the importance of the respective defect for further processing. 
The hierarchy, in order, is as follows: \emph{Rot}, \emph{Discoloration}, \emph{Rot (maybe)}, \emph{Ingrowth/Crack}, \emph{PressureWood}, and \emph{Crosscut}. 
Any remaining annotation artifacts are removed using the morphological operations \emph{remove small holes} and \emph{remove small objects} provided by \emph{scikit-learn} \cite{scikit-learn}.
For experimentation the available training data is split into a \emph{training}, \emph{validation} and test set with a ratio of $0.6, 0.2, 0.2$ respectively.

\section{Computer Vision Models}
\label{sec:models}
\subsection{Intern Image}
\label{sec:sec:intern_image}
InternImage \cite{Wang_2023_CVPR} is a \emph{large scale foundation model} for various vision tasks. It makes use of \emph{deformable convolution v3} (DCNv3) for improved \emph{long range dependencies} and \emph{spatial aggregation}, which is an extension of the original DCN \cite{8237351}. InternImage is based on a transformer-like architecture, using skip connections, feed forward networks and DCNv3 instead of attention. The feature embeddings created using the InternImage model can then be used as input to a variety of computer vision tasks, including semantic segmentation. For this task InternImage provides six pretrained models \emph{tiny (T)} to \emph{huge (H)} of different scales, using the \emph{UPerNet} \cite{upernet} architecture as a segmentation head. For the \emph{H}-scale an additional model using \emph{Mask2Former} \cite{9878483} for segmentation is available.
\subsection{ONE-PEACE}
\label{sec:sec:one_peace}
ONE-PEACE \cite{wang2023one} is a transformer-based \emph{scalable general representation model} for multimodal data. They use a multi-network architecture consisting of \emph{modality adapters}, a central \emph{multi-head self-attention} layer and \emph{modality feed forward networks}. At time of writing ONE-PEACE holds state-of-the-art performance on multiple benchmarks for semantic segmentation, audio-to-text retrieval and image-to-text-retrieval \cite{paperswithcodeOnepeace}. ONE-PEACE provides a pretrained subset model of their main architecture specifically for semantic segmentation using the \emph{Mask2Former} architecture.
\section{Experiments and Results}
\label{sec:experiments}
\subsection{Evaluation Metrics}
\label{sec:sec:evaluation_metrics}
Besides the well established metrics \emph{Accuracy}, \emph{Precision}, \emph{Recall} and \emph{F1-Score}, the \emph{Jaccard coefficient} (IoU) and \emph{Cohen's kappa} are used due to their prominence for semantic segmentation tasks.
Additionally, two custom metrics are created for the specific task of defect detection on logs.
The \emph{ModelScore} is a weighted sum of the class-wise and average F1-Scores of the models, calculated as
\begin{equation}
    ModelScore = \frac{F1_{All} + 2 \times F1_R + F1_{IC}}{4}
\end{equation}
with \emph{R} and \emph{IC} denoting \emph{Rot} and \emph{Ingrowth/Crack} respectively. This metric is used as the primary score on which to rate the model performance during experimentation. The additional weighting of \emph{Rot} and \emph{Ingrowth/Crack} aims to represent the actual defect severity in further processing.
Furthermore, a variation of the \emph{Total Error} (TE) named \emph{PixelDiff} (short: PDiff) is employed. The PixelDiff represents the \emph{absolute percentage of the log's crosscut which is misclassified}. It is calculated as 
\begin{align}
    & PixelDiff = \frac{TE}{\sum (non-background \ classes)} = \frac{|FP| + |FN|}{\sum_{k=1}^{K} (|TP_k| + |FN_k|)} \\
    & PixelDiff_j = \frac{|FP_j| + |FN_j|}{\sum_{k=1}^{K} (|TP_k| + |FN_k|)}
\end{align}
with $K$ being the number of individual classes and $j$ denoting the target class for class-wise metric calculation. The index of $k$ starts at $1$, excluding the \emph{Background} class which has index $0$.
\subsection{Initial Experiments}
\label{sec:sec:initial_experiments}
For initial experimentation, the 6 pretrained semantic segmentation models provided by InternImage, as well as the \emph{Vision Branch} provided by ONE-PEACE are used. Each model is fine-tuned using its default configuration. Tab. \ref{tab:initial_eval} shows the evaluation metrics of the fine-tuned models. The bold columns mark the best performing models of each architecture by ModelScore.
\begin{table}[]
    \centering
    %\scriptsize
    \setlength{\tabcolsep}{5pt}
    \renewcommand{\arraystretch}{1}
    \begin{tabular}{ll|r|r|r|r|r|r|>{\bfseries}r|>{\bfseries}r}
        \toprule
         &  & \multicolumn{7}{l|}{InternImage} & \makecell{ONE-\\PEACE} \\
         \cline{3-10}
         &  & t+uper & s+uper & b+uper & l+uper & xl+uper & h+uper & h+m2f & m2f \\
        \midrule
        \multirow[t]{8}{*}{F1} & All & 0.78 & 0.79 & 0.79 & 0.79 & 0.79 & 0.80 & 0.81 & 0.78 \\
         & BG & 1.00 & 1.00 & 1.00 & 1.00 & 1.00 & 1.00 & 1.00 & 1.00 \\
         & CC & 0.95 & 0.94 & 0.95 & 0.95 & 0.95 & 0.95 & 0.95 & 0.94 \\
         & R & 0.63 & 0.64 & 0.65 & 0.69 & 0.64 & 0.66 & 0.69 & 0.58 \\
         & R(m) & 0.42 & 0.47 & 0.47 & 0.44 & 0.44 & 0.40 & 0.47 & 0.31 \\
         & PW & 0.56 & 0.56 & 0.57 & 0.59 & 0.55 & 0.59 & 0.65 & 0.56 \\
         & DC & 0.78 & 0.73 & 0.77 & 0.79 & 0.77 & 0.76 & 0.78 & 0.75 \\
         & IC & 0.52 & 0.55 & 0.57 & 0.56 & 0.57 & 0.58 & 0.57 & 0.52 \\
        \cline{1-10}
        \multirow[t]{8}{*}{IoU} & All & 0.54 & 0.54 & 0.56 & 0.57 & 0.56 & 0.56 & 0.58 & 0.53 \\
         & BG & 1.00 & 1.00 & 0.99 & 1.00 & 1.00 & 1.00 & 1.00 & 1.00 \\
         & CC & 0.90 & 0.90 & 0.90 & 0.91 & 0.91 & 0.91 & 0.91 & 0.90 \\
         & R & 0.58 & 0.59 & 0.60 & 0.64 & 0.59 & 0.61 & 0.64 & 0.52 \\
         & R(m) & 0.41 & 0.46 & 0.45 & 0.43 & 0.43 & 0.38 & 0.45 & 0.29 \\
         & PW & 0.52 & 0.51 & 0.52 & 0.55 & 0.50 & 0.53 & 0.60 & 0.50 \\
         & DC & 0.74 & 0.69 & 0.74 & 0.76 & 0.73 & 0.72 & 0.74 & 0.71 \\
         & IC & 0.43 & 0.46 & 0.48 & 0.47 & 0.48 & 0.48 & 0.47 & 0.42 \\
        \cline{1-10}
        \multirow[t]{8}{*}{Kappa} & All & 0.94 & 0.94 & 0.94 & 0.94 & 0.94 & 0.95 & 0.94 & 0.94 \\
         & BG & 0.99 & 0.99 & 0.99 & 0.99 & 0.99 & 0.99 & 0.99 & 0.99 \\
         & CC & 0.94 & 0.93 & 0.94 & 0.94 & 0.94 & 0.94 & 0.94 & 0.93 \\
         & R & 0.44 & 0.44 & 0.48 & 0.54 & 0.46 & 0.50 & 0.53 & 0.34 \\
         & R(m) & 0.00 & 0.09 & 0.08 & 0.07 & 0.03 & -0.02 & 0.11 & -0.10 \\
         & PW & 0.29 & 0.28 & 0.28 & 0.35 & 0.28 & 0.33 & 0.44 & 0.26 \\
         & DC & 0.65 & 0.57 & 0.66 & 0.68 & 0.64 & 0.62 & 0.67 & 0.60 \\
         & IC & 0.36 & 0.44 & 0.45 & 0.43 & 0.44 & 0.44 & 0.43 & 0.34 \\
        \cline{1-10}
        \multirow[t]{8}{*}{PDiff} & All & 0.10 & 0.10 & 0.10 & 0.09 & 0.09 & 0.09 & 0.09 & 0.10 \\
         & BG & 0.02 & 0.02 & 0.02 & 0.01 & 0.01 & 0.01 & 0.01 & 0.01 \\
         & CC & 0.08 & 0.09 & 0.08 & 0.08 & 0.08 & 0.08 & 0.08 & 0.08 \\
         & R & 0.21 & 0.22 & 0.19 & 0.17 & 0.20 & 0.18 & 0.18 & 0.25 \\
         & R(m) & 0.43 & 0.38 & 0.40 & 0.39 & 0.42 & 0.43 & 0.37 & 0.43 \\
         & PW & 0.29 & 0.29 & 0.29 & 0.25 & 0.28 & 0.27 & 0.22 & 0.31 \\
         & DC & 0.14 & 0.17 & 0.13 & 0.13 & 0.14 & 0.15 & 0.13 & 0.17 \\
         & IC & 0.16 & 0.11 & 0.12 & 0.14 & 0.14 & 0.15 & 0.15 & 0.19 \\
         \cline{1-10}
        \multicolumn{2}{l|}{ModelScore} & 64.29 & 65.38 & 66.52 & 68.44 & 66.30 & 67.77 & 69.25 & 61.46 \\
        \bottomrule
        \multicolumn{10}{l}{All: mean over all classes, BG: background, CC: Crosscut, R: Rot, R(m): Rot(maybe)} \\
        \multicolumn{10}{l}{PW: PressureWood, DC: Discoloration, IC: Ingrowth/Crack}
    \end{tabular}    
    \caption{Evaluation metrics of the initial experiments}
    \label{tab:initial_eval}
\end{table}
As performing detailed experiments on all available models is neither time- nor cost-effective, 3 models are selected for further experimentation. For InternImage the best performing \emph{InternImage-H-Mask2Former} model is selected. Additionally, while not being the best performing model, \emph{InternImage-H-UPerNet} is also selected for further experimentation. This aims to explore the difference in model performance stemming from the used segmentation head. As ONE-PEACE only provides one pretrained model the \emph{ONE-PEACE-Mask2Former} model is also selected.
\subsection{Casting of Rot(maybe)}
\label{sec:sec:exp_rot_maybe}
\emph{Rot(maybe)} fulfills a special role in defect annotation, as it is the only annotation class not representing its own unique defect type. Instead, \emph{Rot(maybe)} is considered a metaclass made available to annotators to indicate their uncertainty regarding the presence of rot in an area. While it made the annotation process easier for the annotators, determining the true class label for areas marked as \emph{Rot(maybe)} results in a more precise and consistent ground truth for model training.
Analysis of the occurrence of \emph{Rot(maybe)} shows that areas from this class must be attributed to either \emph{Crosscut} or \emph{Rot}.
Wrongful annotations of other defect classes as \emph{Rot(maybe)} are not observed in the dataset. 
To remove \emph{Rot(maybe)} from the dataset an expert is employed to review each annotation featuring \emph{Rot(maybe)}. For this a visualization tool is created, showing both the casting of \emph{Rot(maybe)} to either \emph{Crosscut} or \emph{Rot} for each affected sample. The expert is then tasked with deciding which proposed annotation is correct.
This updated \emph{no\_rm} dataset is then used in training of the 3 selected models. The evaluation metrics for these experiments can be found in Tab. \ref{tab:eval_no_fv}.
The table shows a slight increase in metric scores for the models trained on the updated dataset. As the model objective is to be considered more precise by the removal of \emph{Rot(maybe)}, further experiments are based on the \emph{no\_rm} dataset.
\begin{table}
    \begin{minipage}{.48\linewidth}
    \centering
    %\scriptsize
    \setlength{\tabcolsep}{5pt}
    \renewcommand{\arraystretch}{1}
    \begin{tabular}{ll|r|r|r}
        \toprule
         &  & \multicolumn{2}{r|}{InternImage-h} & \makecell{ONE-\\PEACE} \\
         \cline{3-5}
         &  & uper & m2f & m2f \\
        \midrule
        \multirow[t]{7}{*}{F1} & All & 0.81 & 0.83 & 0.82 \\
         & BG & 1.00 & 1.00 & 1.00 \\
         & CC & 0.95 & 0.95 & 0.95 \\
         & R & 0.65 & 0.68 & 0.55 \\
         & PW & 0.59 & 0.64 & 0.55 \\
         & DC & 0.77 & 0.80 & 0.78 \\
         & IC & 0.59 & 0.61 & 0.52 \\
        \cline{1-5}
        \multirow[t]{7}{*}{IoU} & All & 0.63 & 0.65 & 0.60 \\
         & BG & 1.00 & 1.00 & 1.00 \\
         & CC & 0.92 & 0.92 & 0.91 \\
         & R & 0.59 & 0.62 & 0.49 \\
         & PW & 0.54 & 0.60 & 0.50 \\
         & DC & 0.74 & 0.76 & 0.74 \\
         & IC & 0.49 & 0.51 & 0.42 \\
        \cline{1-5}
        \multirow[t]{7}{*}{Kappa} & All & 0.95 & 0.95 & 0.95 \\
         & BG & 0.99 & 0.99 & 0.99 \\
         & CC & 0.95 & 0.95 & 0.94 \\
         & R & 0.49 & 0.54 & 0.31 \\
         & PW & 0.35 & 0.40 & 0.26 \\
         & DC & 0.65 & 0.69 & 0.65 \\
         & IC & 0.47 & 0.47 & 0.33 \\
        \cline{1-5}
        \multirow[t]{7}{*}{PDiff} & All & 0.08 & 0.08 & 0.08 \\
         & BG & 0.01 & 0.01 & 0.01 \\
         & CC & 0.07 & 0.07 & 0.07 \\
         & R & 0.18 & 0.17 & 0.26 \\
         & PW & 0.25 & 0.25 & 0.31 \\
         & DC & 0.14 & 0.12 & 0.14 \\
         & IC & 0.12 & 0.14 & 0.19 \\
         \cline{1-5}
         \multicolumn{2}{l|}{ModelScore} & 67.44 & 70.13 & 61.18  \\
        \bottomrule
    \end{tabular}    
    \caption{Metrics of models trained and evaluated on the \emph{no\_rm} dataset}
    \label{tab:eval_no_fv}
    \end{minipage}
    \hfill
    \begin{minipage}{.48\linewidth}
    \centering
    %\scriptsize
    \setlength{\tabcolsep}{5pt}
    \renewcommand{\arraystretch}{1}
    \begin{tabular}{ll|r|r|r}
        \toprule
         &  & \multicolumn{2}{r|}{InternImage-h} & \makecell{ONE-\\PEACE} \\
         \cline{3-5}
         &  & uper & m2f & m2f \\
        \midrule
        \multirow[t]{7}{*}{F1} & All & 0.85 & 0.84 & 0.83 \\
         & BG & 1.00 & 1.00 & 1.00 \\
         & CC & 0.97 & 0.96 & 0.96 \\
         & R & 0.75 & 0.72 & 0.58 \\
         & PW & 0.64 & 0.63 & 0.56 \\
         & DC & 0.84 & 0.83 & 0.81 \\
         & IC & 0.68 & 0.64 & 0.58 \\
        \cline{1-5}
        \multirow[t]{7}{*}{IoU} & All & 0.71 & 0.68 & 0.63 \\
         & BG & 1.00 & 1.00 & 1.00 \\
         & CC & 0.94 & 0.94 & 0.93 \\
         & R & 0.70 & 0.66 & 0.52 \\
         & PW & 0.59 & 0.58 & 0.51 \\
         & DC & 0.81 & 0.79 & 0.77 \\
         & IC & 0.59 & 0.54 & 0.48 \\
        \cline{1-5}
        \multirow[t]{7}{*}{Kappa} & All & 0.97 & 0.96 & 0.96 \\
         & BG & 0.99 & 0.99 & 0.99 \\
         & CC & 0.96 & 0.96 & 0.95 \\
         & R & 0.64 & 0.60 & 0.35 \\
         & PW & 0.45 & 0.44 & 0.31 \\
         & DC & 0.76 & 0.74 & 0.72 \\
         & IC & 0.59 & 0.54 & 0.43 \\
        \cline{1-5}
        \multirow[t]{7}{*}{PDiff} & All & 0.05 & 0.06 & 0.06 \\
         & BG & 0.01 & 0.01 & 0.01 \\
         & CC & 0.05 & 0.05 & 0.06 \\
         & R & 0.13 & 0.13 & 0.24 \\
         & PW & 0.20 & 0.20 & 0.27 \\
         & DC & 0.09 & 0.10 & 0.10 \\
         & IC & 0.09 & 0.11 & 0.15 \\
         \cline{1-5}
         \multicolumn{2}{l|}{ModelScore} & 75.95 & 73.12 & 64.14 \\
        \bottomrule
    \end{tabular}    
    \caption{Metrics of models trained and evaluated on the \emph{augmented} dataset}
    \label{tab:eval_augmented}
    \end{minipage}
\end{table}
\subsection{Semi-automatic Ground Truth Correction}
\label{sec:sec:semi-auto_ground-truth}
Visual comparison of the segmentation masks produced by previous experiments to the ground truth annotations shows a huge similarity between the predictions and ground truths for a large portion of the dataset. Visualizing the areas where the predictions differ from the ground truth shows that a significant portion of divergence is made up of only slight deviations in area margins. This suggests that the models may have correctly localized and classified the respective defects, but disagree with the ground truth on the exact area boundaries. 
As the ground truth data is human made, small imperfections and divergences regarding the exact segmentation shape are to be expected.
Therefore, the possibility of the predicted annotation masks being more precise than the ground truth in mapping the segmentation areas' border regions has to be considered.
\newline
To investigate this issue, a comparison similar to the \emph{Rot(maybe)} resolution is created, comparing model-made annotations to the ground truth. An unbiased expert is then tasked with selecting the annotation which in his opinion more precisely matches the actual defects on the logs. To further minimize bias towards either the human-made or predicted annotations, the expert is not informed which of the presented annotations stems from which source. 
The resulting \emph{augmented} dataset is proposed to contain more precise annotations than the original dataset. Similar to the \emph{no\_rm} dataset the 3 selected models are then trained and evaluated using the \emph{augmented} dataset. The evaluation of these models is shown in Tab. \ref{tab:eval_augmented}.
Comparing these results to the previous experiments on the \emph{no\_rm} dataset, the metric scores increased across all target classes for all models, except \emph{PressureWood} on \emph{InternImage-H-Mask2Former}.
A micro-average evaluation shows a significant increase in \emph{true positive rate}, with simultaneous decrease in \emph{false positive rate} across all models and most defect classes.
Furthermore, the high confusion between all defect classes and \emph{Crosscut} observed in previous experiments is reduced significantly. This indicates the shape of the predictions produced by the models trained on the \emph{augmented} datasets more closely matches the updated ground truth, which is the expected effect of using the \emph{augmented} dataset
\subsection{Analysis of the best performing Model}
\label{sec:sec:exp_curr_learning}
\begin{figure}
    \centering
    \includegraphics[scale=0.5]{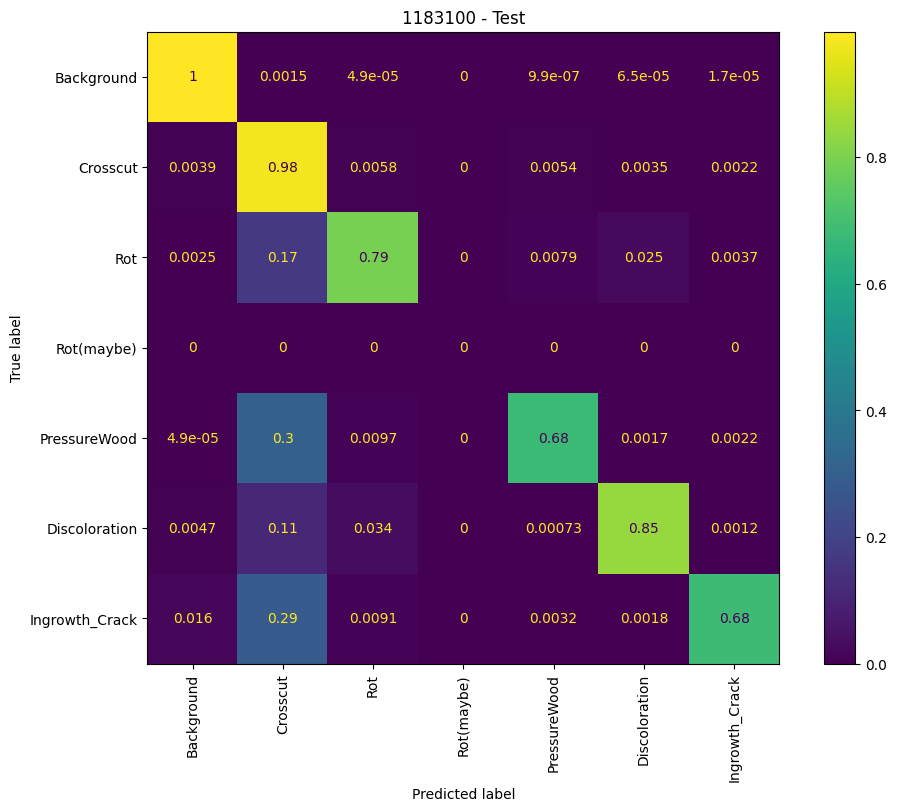}
    \caption{Micro-average confusion matrix of the \emph{InternImage-H-UperNet} model on the \emph{test} dataset}
    \label{fig:best_confmat_test}
\end{figure}

Considering the ModelScore, the best performing model is \emph{InternImage-H-UperNet-augmented} with a ModelScore of $75.95\%$. The detailed evaluation metrics for this model are already shown in Tab. \ref{tab:eval_augmented}. Fig. \ref{fig:best_confmat_test} shows the micro-average confusion matrix for the model on the test dataset.

\begin{figure}
    \centering
    \includegraphics[width=1\linewidth]{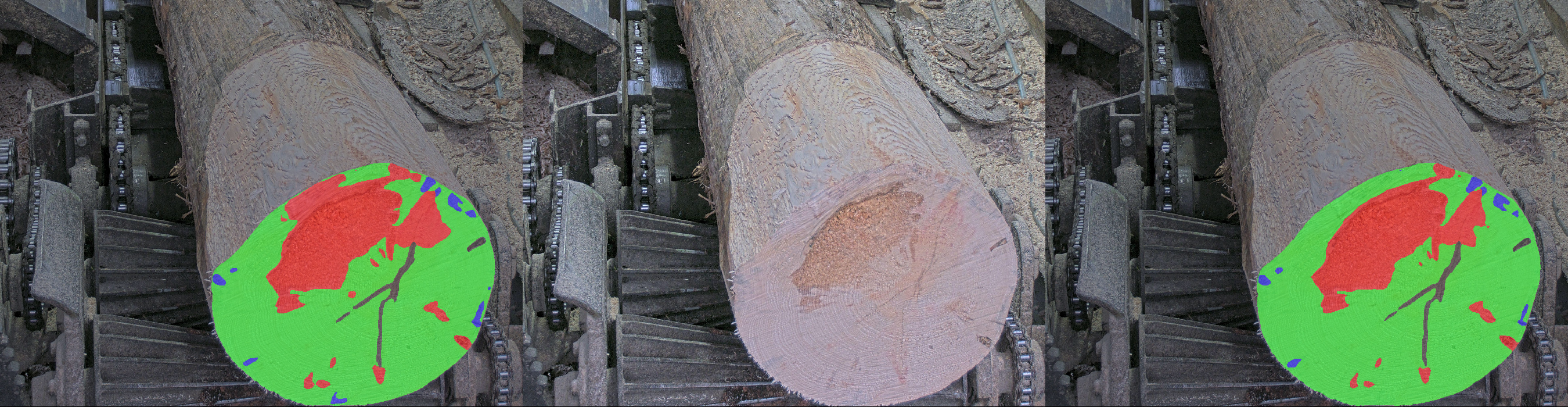}
    \caption{Comparison of the ground truth annotation (left) and prediction of the best model (right) to the original image (center)}
    \label{fig:mask_comparison_example}
\end{figure}

The confusion matrix shows virtually no false positives or false negatives between defect classes. A large misclassification is observed in the form of \emph{false negatives} between the defect classes and the \emph{Crosscut} class. 
Visual inspection of the predicted annotation masks confirms that this is the result of annotation inaccuracy introduced by the human annotators, as well as wrong or debatable ground truth annotations remaining after the correction. 
\begin{figure}
    \centering
    \includegraphics[width=1\linewidth]{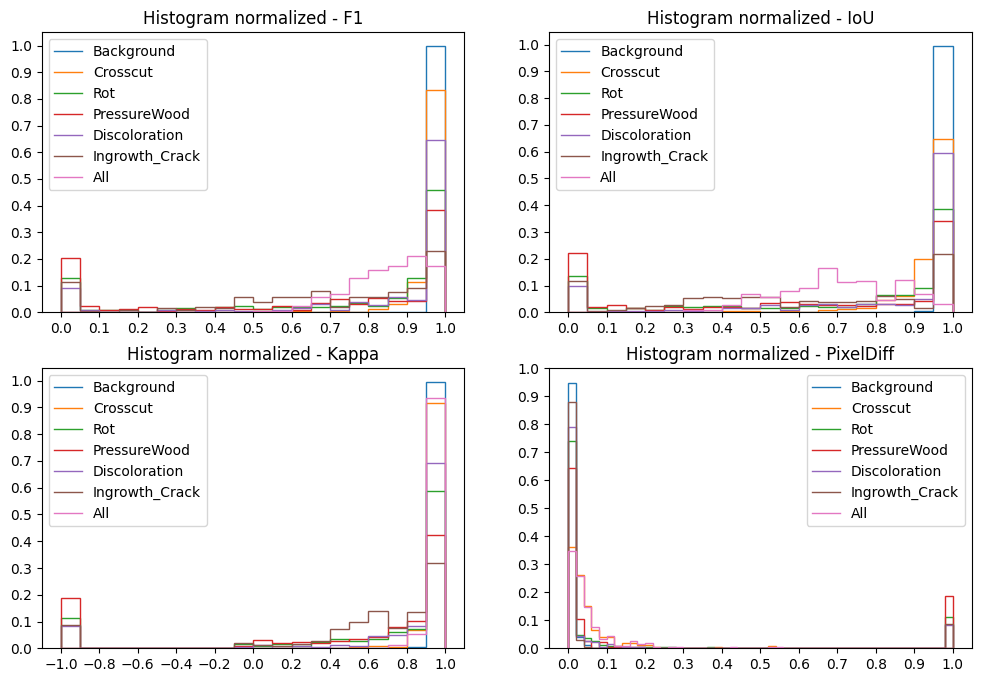}
    \caption{Histograms of the sample distribution for the respective metrics on the test set}
    \label{fig:best_hists}
\end{figure}
%
%\newline
Fig. \ref{fig:best_hists} shows the sample distribution across the range of the respective evaluation metrics. These show that the model performs reasonably well for a majority of the test samples. Notably a significant portion of samples is either ranked with the best or worst possible metric score for the class-wise metrics. This is due to the edge case handling required for samples where either or both ground truth and prediction do not feature a specific target class. If both ground truth and prediction do not contain instances of a class, the metric is by design set to the best possible value. Similarly, when only one of the two annotations features the specific class, the metric for this class is set to the worst possible value.
%
%\newline
%\newline
%
Visually inspecting the produced segmentation masks shows a large similarity between the predictions and ground truth for a majority of samples. An example for a segmentation produced by the best model is shown in Fig. \ref{fig:mask_comparison_example}.
\subsection{Inter-Annotator Agreement Analysis}
\label{sec:sec:inter_rater}
Due to the highly subjective nature of our annotation task and the varying knowledge levels of the partaking annotators, great variance within the ground truth annotations is expected.
This is already hinted at by the dataset correction rate of \(33.7\%\).
We add our \emph{best-model} as an additional annotator to compare its performance against our expert and layman annotators.
Our agreement analysis is performed exclusively on the \emph{warmup} dataset, which was annotated by all five annotators and was withheld from model training, testing and evaluation.
We use statistical measures \emph{Cohen's kappa} \cite{doi:10.1177/001316446002000104,McHugh2012-gs} and the \emph{Jaccard similarity coefficient} (IoU) \cite{jaccard_iou} to measure the agreement between annotators.
\emph{Cohen's kappa} returns scores in the interval \([-1, 1]\), with \(1\) denoting perfect agreement, \(0\) denoting no agreement, and scores below \(0\) denoting a inverse agreement.
The \emph{Jaccard similarity coefficient} returns scores in the interval \([0, 1]\), with \(1\) denoting perfect correlation and \(0\) denoting no correlation.
We use expert annotator \emph{B} as a \emph{baseline} for comparison, both due to their expertise and involvement in the creation of our dataset.
Tab. \ref{tab:inter_rater} shows the class-wise mean agreements measures.
\begin{table}
    \begin{center}
        %\scriptsize
        \renewcommand{\arraystretch}{1}
        \begin{tabular}{llrrrrrrrr}
            \toprule
             & Annotator & All & BG & CC & R & R(m) & PW & DC & IC \\
            \midrule
            \multicolumn{10}{l}{Cohen's kappa} \\
            \midrule
            \multirow[t]{5}{*}{B} & A & 0.951 & 0.989 & 0.936 & 0.809 & 0.480 & 0.709 & 0.709 & -0.097 \\
             & C & 0.928 & 0.982 & 0.921 & 0.464 & 0.205 & 0.023 & 0.511 & 0.094 \\
             & D & 0.913 & 0.980 & 0.903 & 0.736 & -0.141 & 0.209 & 0.653 & 0.263 \\
             & E & 0.930 & 0.985 & 0.925 & 0.626 & -0.016 & 0.410 & 0.601 & -0.185 \\
             & best-model & 0.944 & 0.987 & 0.936 & 0.612 & 0.480 & 0.151 & 0.540 & 0.340 \\
            \midrule
            \multicolumn{10}{l}{Jaccard coefficient (IoU)} \\
            \midrule
            \multirow[t]{5}{*}{B} & A & 0.623 & 0.993 & 0.984 & 0.807 & 0.740 & 0.801 & 0.803 & 0.197 \\
             & C & 0.518 & 0.990 & 0.976 & 0.648 & 0.561 & 0.400 & 0.679 & 0.23 \\
             & D & 0.541 & 0.988 & 0.973 & 0.780 & 0.370 & 0.515 & 0.767 & 0.3131 \\
             & E & 0.534 & 0.991 & 0.979 & 0.731 & 0.441 & 0.647 & 0.765 & 0.180 \\
             & best-model & 0.548 & 0.991 & 0.978 & 0.615 & 0.740 & 0.433 & 0.635 & 0.294 \\
            \bottomrule
            \multicolumn{10}{l}{All: mean over all classes, BG: background, CC: Crosscut, R: Rot, R(m): Rot(maybe)} \\
            \multicolumn{10}{l}{PW: PressureWood, DC: Discoloration, IC: Ingrowth/Crack}
            \end{tabular}            
        \caption{Class-wise mean \emph{Cohen's kappa} and \emph{Jaccard similarity coefficient} between our annotators \emph{B} to \emph{E} our \emph{best-model} and our \emph{ground truth} annotator \emph{A}}
        \label{tab:inter_rater}
    \end{center}
\end{table}
The table in general shows higher agreement between the expert annotators than in between expertise levels.
It also shows that despite possessing similar expertise, expert annotators still disagree substantially for some defect classes, which indicates differences in subjective evaluation of the defects.
The best model shows agreement with the baseline annotations comparable to the agreement between experts and laymen. Assuming the correctness of the expert annotations this suggests the performance of the model is at least on par with the laymen annotators.
\section{Conclusions}
We were able to produce segmentation models capable of properly segmenting the images of wooden logs in regard to their defects. Both evaluation metrics and visual examination of the best model show that the model produces segmentation masks very close to the provided ground truth for a majority of data samples. For those samples where the model prediction and ground truth differ significantly, a large uncertainty regarding the correctness of the ground truth can be observed. 
While the model may not perform well enough for fully autonomous utilization, it is already well suited to being used in a production environment, either as a decision-assistance system, or autonomous but with human supervision. 
\section{Future Work}
As the most limiting factor for this project was the small amount of annotation data created by experts, future work may create even better models through the use of more and more consistent annotation data.
Furthermore, as a large amount of time for this project had to be invested into dealing with the limited data, further research may also expend more effort into the hyperparameter configuration of the models, possibly further increasing model performance.
Lastly, while the explored architectures produced sensible results, different model architectures may prove more suitable for this specific task. Therefore, exploring other deep learning architectures may provide valuable insights into the types of architectures best applicable to this task.
\section*{Acknowledgments}
We gratefully acknowledge the scientific support and HPC resources provided by the Erlangen National High Performance Computing Center (NHR@FAU) of the Friedrich-Alexander-Universität Erlangen-Nürnberg (FAU) under the NHR project b196ac14.
We would like to thank Sägewerk Müller-Gei, local Franconian sawmill for their cooperation and insight.
%
%We would like to thank BLIND, REVIEW for their cooperation and insight.
%
%% \bibliography{lni-paper-example-de.tex} ist hier nicht erlaubt: biblatex erwartet dies bei der Preambel
%% Starten Sie "biber paper", um eine Biliographie zu erzeugen.
\pagebreak

\printbibliography

@article{doi:10.1139/cjfr-2020-0340,
author = {Soge, Ayodele O. and Popoola, Olatunde I. and Adetoyinbo, Adedeji A.},
title = {Detection of wood decay and cavities in living trees: a review},
journal = {Canadian Journal of Forest Research},
volume = {51},
number = {7},
pages = {937-947},
year = {2021},
doi = {10.1139/cjfr-2020-0340},
URL = {https://doi.org/10.1139/cjfr-2020-0340},
eprint = {https://doi.org/10.1139/cjfr-2020-0340}
}

@article{dac3411a39e9484e9284ea2ea969adf6,
title = "Automatic detection of root rot and resin in felled Scots pine stems using convolutional neural networks",
author = "Eero Holmstrom and Henna Kainulainen and Antti Raatevaara and Jonne Pohjankukka and Tuula Piri and Juha Honkaniemi and Jori Uusitalo and Mikko Peltoniemi and Aleksi Lehtonen",
year = "2024",
month = mar,
day = "14",
doi = "10.1080/14942119.2024.2327247",
language = "English",
journal = "International Journal of Forest Engineering",
issn = "1494-2119",
publisher = "Faculty of Forestry & Environmental Management, University of New Brunswick",
url = {https://www.tandfonline.com/doi/full/10.1080/14942119.2024.2327247}
}

@article{s19071579,
AUTHOR = {Ostovar, Ahmad and Talbot, Bruce and Puliti, Stefano and Astrup, Rasmus and Ringdahl, Ola},
TITLE = {Detection and Classification of Root and Butt-Rot (RBR) in Stumps of Norway Spruce Using RGB Images and Machine Learning},
JOURNAL = {Sensors},
VOLUME = {19},
YEAR = {2019},
NUMBER = {7},
ARTICLE-NUMBER = {1579},
URL = {https://www.mdpi.com/1424-8220/19/7/1579},
PubMedID = {30939827},
ISSN = {1424-8220},
DOI = {10.3390/s19071579}
}

@article{HE2020107357,
title = {Application of deep convolutional neural network on feature extraction and detection of wood defects},
journal = {Measurement},
volume = {152},
pages = {107357},
year = {2020},
issn = {0263-2241},
doi = {https://doi.org/10.1016/j.measurement.2019.107357},
url = {https://www.sciencedirect.com/science/article/pii/S0263224119312217},
author = {Ting He and Ying Liu and Yabin Yu and Qian Zhao and Zhongkang Hu}
}

@article{hrcak278445,
  author          = {Haciefendioglu, Kemal and Basri Başaga, Hasan and Emre Kartal, Murat and Ceyhun Bulut, Mehmet},
  journal         = {Drvna industrija},
  number          = {2},
  title           = {Automatic Damage Detection on Traditional Wooden Structures with Deep Learning-Based Image Classification Method},
  volume          = {73},
  pages           = {163-176},
  year            = {2022},
  doi             = {https://doi.org/10.5552/drvind.2022.2108},
  url             = {https://hrcak.srce.hr/file/403148},
}

@InProceedings{10.1007/978-3-031-51026-7_2,
author="Baroni, Giulia L.
and Rasotto, Laura
and Roitero, Kevin
and Siraj, Ameer Hamza
and Della Mea, Vincenzo",
editor="Foresti, Gian Luca
and Fusiello, Andrea
and Hancock, Edwin",
title="Vision Transformers for Breast Cancer Histology Image Classification",
booktitle="Image Analysis and Processing - ICIAP 2023 Workshops",
year="2024",
publisher="Springer Nature Switzerland",
address="Cham",
pages="15--26",
isbn="978-3-031-51026-7"
}

@inproceedings{dosovitskiy2021an,
title={An Image is Worth 16x16 Words: Transformers for Image Recognition at Scale},
author={Alexey Dosovitskiy and Lucas Beyer and Alexander Kolesnikov and Dirk Weissenborn and Xiaohua Zhai and Thomas Unterthiner and Mostafa Dehghani and Matthias Minderer and Georg Heigold and Sylvain Gelly and Jakob Uszkoreit and Neil Houlsby},
booktitle={International Conference on Learning Representations},
year={2021},
url={https://openreview.net/forum?id=YicbFdNTTy}
}

@article{KER2019239,
title = {Automated brain histology classification using machine learning},
journal = {Journal of Clinical Neuroscience},
volume = {66},
pages = {239-245},
year = {2019},
issn = {0967-5868},
doi = {https://doi.org/10.1016/j.jocn.2019.05.019},
url = {https://www.sciencedirect.com/science/article/pii/S0967586819306563},
author = {Justin Ker and Yeqi Bai and Hwei Yee Lee and Jai Rao and Lipo Wang}
}

@INPROCEEDINGS{7298594,
  author={Szegedy, Christian and Wei Liu and Yangqing Jia and Sermanet, Pierre and Reed, Scott and Anguelov, Dragomir and Erhan, Dumitru and Vanhoucke, Vincent and Rabinovich, Andrew},
  booktitle={2015 IEEE Conference on Computer Vision and Pattern Recognition (CVPR)}, 
  title={Going deeper with convolutions}, 
  year={2015},
  volume={},
  number={},
  pages={1-9},
  doi={10.1109/CVPR.2015.7298594},
  ISSN={1063-6919},
  month= jun
}

@article{ScheiklLaschewski2020,
url = {https://doi.org/10.1515/cdbme-2020-0016},
title = {Deep learning for semantic segmentation of organs and tissues in laparoscopic surgery},
author = {Paul Maria Scheikl and Stefan Laschewski and Anna Kisilenko and Tornike Davitashvili and Benjamin Müller and Manuela Capek and Beat P. Müller-Stich and Martin Wagner and Franziska Mathis-Ullrich},
pages = {20200016},
volume = {6},
number = {1},
journal = {Current Directions in Biomedical Engineering},
doi = {doi:10.1515/cdbme-2020-0016},
year = {2020},
lastchecked = {2024-04-16}
}

@article{LWFFichte,
  author       = {Cornelia Triebenbacher and Ludwig Straßer and Ralf Petercord},
  title   = {Waldschutzrisiko der Fichte},
  journal = {LWF Wissen},
  volume = {80},
  url        = {https://www.lwf.bayern.de/waldschutz/monitoring/171673/index.php},
  year         = {2017},
  month        = sep,
}

@misc{LabelStudio,
	title={{{Label Studio}: Data labeling software}},
	url={https://github.com/heartexlabs/label-studio},
	author={Maxim Tkachenko and Mikhail Malyuk and Andrey Holmanyuk and Nikolai Liubimov},
	year={2020-2022},
}

@inproceedings{SAM,
  title={Segment anything},
  author={Kirillov, Alexander and Mintun, Eric and Ravi, Nikhila and Mao, Hanzi and Rolland, Chloe and Gustafson, Laura and Xiao, Tete and Whitehead, Spencer and Berg, Alexander C and Lo, Wan-Yen and others},
  booktitle={Proceedings of the IEEE/CVF International Conference on Computer Vision},
  pages={4015--4026},
  year={2023}
}

@InProceedings{Wang_2023_CVPR,
    author    = {Wang, Wenhai and Dai, Jifeng and Chen, Zhe and Huang, Zhenhang and Li, Zhiqi and Zhu, Xizhou and Hu, Xiaowei and Lu, Tong and Lu, Lewei and Li, Hongsheng and Wang, Xiaogang and Qiao, Yu},
    title     = {InternImage: Exploring Large-Scale Vision Foundation Models With Deformable Convolutions},
    booktitle = {Proceedings of the IEEE/CVF Conference on Computer Vision and Pattern Recognition (CVPR)},
    month     = {June},
    year      = {2023},
    pages     = {14408-14419}
}

@INPROCEEDINGS{8237351,
  author={Dai, Jifeng and Qi, Haozhi and Xiong, Yuwen and Li, Yi and Zhang, Guodong and Hu, Han and Wei, Yichen},
  booktitle={2017 IEEE International Conference on Computer Vision (ICCV)}, 
  title={Deformable Convolutional Networks}, 
  year={2017},
  pages={764-773},
  doi={10.1109/ICCV.2017.89},
  ISSN={2380-7504},
  month={Oct},
}

@INPROCEEDINGS{9878483,
  author={Cheng, Bowen and Misra, Ishan and Schwing, Alexander G. and Kirillov, Alexander and Girdhar, Rohit},
  booktitle={2022 IEEE/CVF Conference on Computer Vision and Pattern Recognition (CVPR)}, 
  title={Masked-attention Mask Transformer for Universal Image Segmentation}, 
  year={2022},
  pages={1280-1289},
  doi={10.1109/CVPR52688.2022.00135},
  ISSN={2575-7075},
  month={June},}

@InProceedings{upernet,
author="Xiao, Tete
and Liu, Yingcheng
and Zhou, Bolei
and Jiang, Yuning
and Sun, Jian",
editor="Ferrari, Vittorio
and Hebert, Martial
and Sminchisescu, Cristian
and Weiss, Yair",
title="Unified Perceptual Parsing for Scene Understanding",
booktitle="Computer Vision -- ECCV 2018",
year="2018",
publisher="Springer International Publishing",
address="Cham",
pages="432--448",
isbn="978-3-030-01228-1"
}

@article{wang2023one,
  title={ONE-PEACE: Exploring One General Representation Model Toward Unlimited Modalities},
  author={Wang, Peng and Wang, Shijie and Lin, Junyang and Bai, Shuai and Zhou, Xiaohuan and Zhou, Jingren and Wang, Xinggang and Zhou, Chang},
  journal={arXiv preprint arXiv:2305.11172},
  note={Submitted to ICLR 2024. Status: Rejected (2024.02.16) \url{https://openreview.net/forum?id=9Klj7QG0NO}.},
  year={2023}
}

@misc{paperswithcodeOnepeace,
  author       = {{\relax{Meta AI Research}}},
  title        = {Papers With Code: ONE-PEACE},
  url          = {https://paperswithcode.com/paper/one-peace-exploring-one-general},
  howpublished = {\url{https://paperswithcode.com/paper/one-peace-exploring-one-general}},
  note         = {accessed: 2024.05.02},
  year         = {2024}
}

@article{scikit-learn,
 title={Scikit-learn: Machine Learning in {P}ython},
 author={Pedregosa, F. and Varoquaux, G. and Gramfort, A. and Michel, V.
         and Thirion, B. and Grisel, O. and Blondel, M. and Prettenhofer, P.
         and Weiss, R. and Dubourg, V. and Vanderplas, J. and Passos, A. and
         Cournapeau, D. and Brucher, M. and Perrot, M. and Duchesnay, E.},
 journal={Journal of Machine Learning Research},
 volume={12},
 pages={2825--2830},
 year={2011}
}

@article{doi:10.1177/001316446002000104,
author = {Jacob Cohen},
title ={A Coefficient of Agreement for Nominal Scales},
journal = {Educational and Psychological Measurement},
volume = {20},
number = {1},
pages = {37-46},
year = {1960},
doi = {10.1177/001316446002000104},
URL = {https://doi.org/10.1177/001316446002000104},
}

@ARTICLE{McHugh2012-gs,
  title    = "Interrater reliability: the kappa statistic",
  author   = "McHugh, Mary L",
  journal  = "Biochem Med (Zagreb)",
  volume   =  22,
  number   =  3,
  pages    = "276--282",
  year     =  2012
}

@article{jaccard_iou,
author = {Jaccard, Paul},
year = {1902},
pages = {69-130},
title = {Lois de distribution florale dans la zone alpine},
volume = {38},
journal = {Bulletin de la Société vaudoise des Sciences Naturelles},
doi = {10.5169/seals-266762}
}
\end{document}